\documentclass[sn-mathphys-ay]{sn-jnl}
\usepackage{graphicx} 
\usepackage{xcolor}
\usepackage{booktabs}
\usepackage{url}
\usepackage{multirow}

\usepackage{graphicx}
\usepackage{adjustbox}
\usepackage{tablefootnote}
\usepackage{amsmath,amssymb,amsfonts}%
\usepackage{amsthm}%
\usepackage{mathrsfs}%
\usepackage[title]{appendix}%
\usepackage{xcolor}%
\usepackage{textcomp}%
\usepackage{manyfoot}%

\def \vardatumod{4/2/2023}
\def \vardatumdo{20/5/2023}
\def \twitterklicovaslova{Ukraine, Russia, Zelensky, Putin}

\def \twitterpocetcelkem{34\,124}

\begin{document}

\title{Fine-tuning multilingual language models in Twitter/X sentiment analysis: a study on Eastern-European V4 languages}
%
%

\author[1]{\fnm{Tom\'a\v s} \sur{Filip}}\email{F150520@fpf.slu.cz}
\equalcont{These authors contributed equally to this work.}

\author[1]{\fnm{Martin} \sur{Pavl\'\i\v cek}}\email{martin.pavlicek@fpf.slu.cz}
\equalcont{These authors contributed equally to this work.}

\author*[1,2]{\fnm{Petr} \sur{Sos{\'\i}k}}\email{Petr.Sosik@osu.cz}

\affil[1]{\orgdiv{Institute of Computer Science}, \orgname{Faculty of Philosophy and Science, Silesian University in Opava}, \orgaddress{\street{Bezru\v covo n\'am. 13}, \city{Opava}, \postcode{74601}, \country{Czech Republic}}}

\affil[2]{\orgdiv{Institute for Research and Applications of Fuzzy Modeling}, \orgname{University of Ostrava}, \orgaddress{\street{30. dubna 22}, \city{Ostrava}, \postcode{70200}, \country{Czech Republic}}}

\abstract{
The aspect-based sentiment analysis (ABSA) is a standard NLP task with numerous approaches and benchmarks, where large language models (LLM) represent the current state-of-the-art. We focus on ABSA subtasks based on Twitter/X data in underrepresented languages. On such narrow tasks, small tuned language models can often outperform universal large ones, providing available and cheap solutions.

We fine-tune several LLMs (BERT, BERTweet, Llama2, Llama3, Mistral)  for classification of sentiment towards Russia and Ukraine in the context of the ongoing military conflict. The training/testing dataset was obtained from the academic API from Twitter/X during 2023, narrowed to the languages of the V4 countries (Czech Republic, Slovakia, Poland, Hungary). Then we measure their performance under a variety of settings including translations, sentiment targets, in-context learning and more, using GPT4 as a reference model. We document several interesting phenomena demonstrating, among others, that some models are much better fine-tunable on multilingual Twitter tasks than others, and that they can reach the SOTA level with a very small training set. Finally we identify combinations of settings providing the best results. 
}

\maketitle


\section{Introduction}
\label{sec_intro}
Aspect-based sentiment analysis (ABSA) \citep{nazir2020issues} includes a collection of methods extracting sentiment towards a specific aspect associated with a given target in a message. Usual sources are text documents of various lengths, although other media (speech, video etc.) are included in multimodal sentiment analysis \citep{kaur2022multimodal}. Methods of automated sentiment identification can be classified into into three major categories: knowledge-based, machine learning, and hybrid models \citep{brauwers2022survey}. 

While machine learning methods, especially attention-based models as transformers, are considered the current state-of-the-art\footnote{https://paperswithcode.com/task/aspect-based-sentiment-analysis} on sufficiently large training datasets (see, e.g. \citep{scaria2023instructabsa}), the lack of training data (especially in languages other than English) can decrease their efficiency significantly. In these cases, either simpler machine-learning methods such as SVQ's, or hybrid methods augmenting machine-learning models with knowledge bases and structures can be the choice. Another solution may be a transfer of knowledge from other languages where sufficiently large databases exist \citep{barbieri2022xlm}.

ABSA can be described in three phases \citep{brauwers2022survey}: aspect detection/extraction, sentiment classification, and sentiment aggregation. In this paper we use Twitter/X as the main (and rapid) source of public sentiment data, where the aspect detection is usually simple due to their restricted length and the possibility to follow re-tweets forming a thread dealing with a certain aspect/target. The sentiment classification, however, can be challenging: when tweets are processed alone, we see a lack of context, all the more so since the tweets relate to contemporary topics and assume the reader's knowledge of the current context. Therefore, even large universal models as GPT4 struggle with their specific style, and smaller fine-tuned models often perform better, especially in the case of non-English datasets. This fact is confirmed in several benchmarks and also in our experiments.

We focus on the analysis of sentiment towards the mostly publicised world event of the last few years: the Ukraine war crisis. The conflict even before it's kinetic beginning was waged in cyberspace with methods of soft (information and narrative dissemination on social networks) and hard (ransomware, theft, DDoS) cyber attacks \citep{ducheine_next_2022}. Hence, a large online textual corpus is available for training and testing. 
There exist numerous studies devoted to ABSA in tweets, including those focused on the recent Russo-Ukraine conflict. Nevertheless, we are not aware of any LLM-based sentiment analysis on this topic in Eastern European languages (the V4 group), although these countries form the previous Russian sphere of influence which is now openly claimed back by Kremlin, and are subject of intensive cyber-bullying attacks from Russia. Neither we have met a comparison of performance of LLama and Mistral models to the BERT-derived ones on the task of multilingual tweet analysis similar to ours. 

\begin{figure}[ht!]
    \centering
    \includegraphics[width=\linewidth]{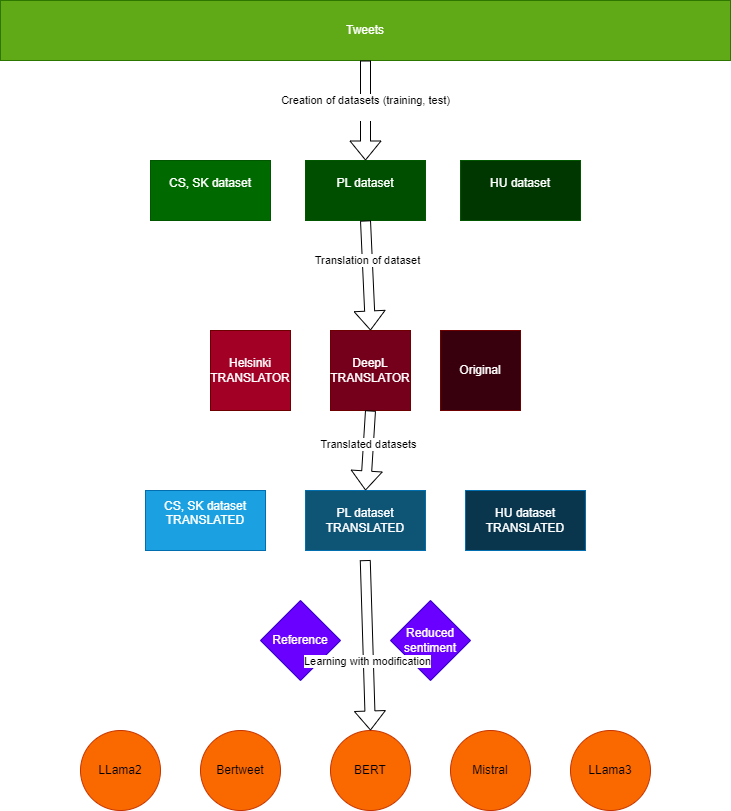}
    \caption{Illustration of the experimental pipeline. The downloaded dataset was split into three language-specific parts which were annotated. Three version versions of translation (Helsinki, DeepL, none) were prepared, obtaining 9 individual datasets. Finally, the models were fine-tuned and tested. Experiments were run in four variants, combining classification into two/three classes and training with/without reference tweets. }
    \label{fig:enter-label}
\end{figure}
   
Our experimental pipeline started by downloading data from Twitter/X academic API during 2023, filtered by keywords and restricted to the V4 languages (Czech Republic, Slovakia, Poland, Hungary). The obtained language-specific datasets were manually labelled. Apart from the original datasets, we also produced their versions translated to English using the Helsinki translator and DeepL. Then we fine-tuned the following models: BERT, BERTweet, Llama2, Llama3, Mistral with a training part of our datasets and then measured standard metrics (accuracy, recall, precision, F1) using the testing parts. The training objective was the sentiment polarity towards either Ukraine or Russia (negative/positive/neutral). We used many combinations of setting: two/three polarities classification, usage of the reference tweet, in-context learning and more. We also employed GPT4 as the reference model (without fine-tuning). The main findings are summarised as follows: 
\begin{enumerate}
    \item[(a)] fine-tuning with as few as 6K multilingual tweets provided significantly better (SOTA level) results  than in-context learning; 
    \item[(b)] the performance of the tested models on our Twitter/X corpus was often uncorrelated  with their results in general benchmarks; 
    \item[(c)] a good translation to English provided an advantage over the use of the original languages even for multilingual pre-trained models; 
    \item[(d)] some models showed unexpected language- and culture-specific differences arising from a wider context.
\end{enumerate}


\section{Background}
\label{sec_background}

A recent review of results in NLP-based sentiment analysis can be found in \citep{jim2024recent}. 
Performance assessment of models analysing sentiment in tweets was covered in TweetEval \citep{barbieri2020tweeteval}, summarising datasets and methodology to conduct benchmarks in tasks of detection of emotion, irony, hate speech, offensive language, stance, emoji prediction and sentiment analysis. The tasks were described in the proceedings of series of the International Workshops on Semantic Evaluation, \url{https://semeval.github.io/}.

The TweetEval leaderboard at GitHub lists BERTweet \citep{nguyen2020bertweet} as the current SoTA model, although closely followed by TimeLM-21. The family of TimeLM models  \citep{loureiro2022timelms} reflects the current context problem by periodical updates with tweet datasets and it showed excellent results on up-to-date topics, outperforming BERTweet in many tasks.

Extending focus on multilingual tweet analysis, \citep{barbieri2022xlm} presented a unified dataset of tweets in eight languages for benchmarking (UMSAB). The paper also introduced the model XLM-Twitter (XLM-T) developed by pre-training the XLM-R \citep{conneau2020unsupervised} from publicly available checkpoints using 198M multilingual tweets dated from 2018 to 2020. XLM-R itself is a multilingual extension of RoBERTa \citep{liu2019roberta}. XLM-T was then fine-tuned and tested on UMSAB, and while it slightly underperformed monolingual models as TimeLMs or BERTweet, it can be considered the SoTA among multilingual Twitter-specific models.

Another study \citep{barreto2023sentiment} focused on (i) effectiveness of various static embeddings for sentiment analysis in tweets; (ii) performance of BERT, RoBERTa and BERTweet in Twitter ABSC tasks, confirming BERTweet as SoTA; (iii) further specific pre-training of these models to improve their performance. 

Finally, there exist numerous studies on Russia--Ukraine war sentiment and emotion analysis on social networks, using mostly either BERT-based models or non-neural ML approaches. We mention just two examples. An evaluation of traditional supervised machine learning models (logistic regression, decision trees, random forests, SVMs and others) on Twitter data was provided in \citep{wadhwani2023sentiment}. The paper can serve as a baseline to our results as it used the same kind of data and ABSC tasks. The evaluation was based on 11,250 English tweets and the best performing model reported accuracy of 0.84. A deep learning approach to analysis of tweets about the Russia Ukraine conflict was presented in \citep{aslan2023deep} using a combination of multi-feature CNN with BiLSTM and the FastText embedding. Both these studies relied on monolingual English datasets.


\section{Methods}
\label{sec_methods}
\subsection*{Data collection}
 
Data were collected using the academic Twitter/X API during the period  \vardatumod\ to \vardatumdo. We set the filters as follows: (i) languages -- Czech, Slovak, Polish, Hungarian; (ii) keywords \twitterklicovaslova{} representing the topic we wanted to focus on. The total number of collected tweets was \twitterpocetcelkem.

The dataset was then split into three subsets based on the original language Czech/Slovak (CS), Polish (PL) and Hungarian (HU). We could not create a separate Slovak dataset  as there was no filter available for Slovak so it was mixed with Czech, and an automated Google language recognition we tried was unable to distinguish reliably between these two close languages.

\subsection*{Data annotation}
From every language-specific subset we randomly chose a certain part for annotation. Tweets were manually annotated due to their sentiment (1-negative, 2-neutral, 3-positive) towards Ukraine or Russia. The annotation intended to provide roughly equal number of tweets in all categories in each language and it did not reflect the overall ratio of positive, negative and neutral tweets in that language. These ratios varied, e.g., the Hungarian subset contained lower ratio of positive tweets  towards Ukraine than the others. We tried also machine labelling using the pre-trained model GPT4 but the results were inconclusive. 
Each annotated subset was further split into a training part (75 \%) and testing part (25 \%). The sizes of all annotated sets are listed in Table \ref{tab:training}.

\subsection*{Translation}
The annotated datasets were used for both training and testing in three different language modes: 
\begin{itemize}
    \item translated to English using the Helsinki Neural Machine Translation System\footnote{https://huggingface.co/Helsinki-NLP};
    \item translated to English using the DeepL API\footnote{https://www.deepl.com/translator};
    \item no translation, original languages (CS/SK, PL, HU).
\end{itemize}

\subsection*{Language models}

The following table lists the models used in our experiments. Furthermore, GPT4 \citep{GPT4-report} was used as a reference model for tweet classification. As fine tuning was not available at the time of the experiments, instead we used in-context learning described bellow.

\begin{table}[h!]
    \centering
    \caption{Language models used in the experiments. }
    \label{tab:models}
    \begin{tabular}{llll}
    \toprule
      \bf Model & \bf Total  & \bf Fine-tuned  & \bf Reference\\
                & \bf params & \bf params & \bf \\
         \midrule
BERT base-uncased\tablefootnote{https://huggingface.co/google-bert/bert-base-uncased} & 110M & 110M & \citep{devlin2019pre} \\
BERTweet large\tablefootnote{https://huggingface.co/vinai/bertweet-large} & 355M & 355M & \citep{nguyen2020bertweet}\\
LLama-2 7B\tablefootnote{https://huggingface.co/meta-llama/Llama-2-7b-hf} & 6.7B & 4M &\citep{touvron2023llama}\\
LLama-3 8B\tablefootnote{https://huggingface.co/meta-llama/Meta-Llama-3-8B} & 8B & 3.4M & \citep{llama3modelcard}\\
Mistral 7B\tablefootnote{https://huggingface.co/mistralai/Mistral-7B-v0.1} & 7.2B & 3.4M & \citep{jiang2023mistral}\\
         \botrule
    \end{tabular}
\end{table}

\subsection*{Training}
The experiments included models with sufficient capabilities (e.g., BERTweet is declared the SOTA model for tweet sentiment analysis at the TweetEval benchmark), but simultaneously up to billions of parameters, so that they could be run on a limited GPU hardware. 

\begin{table}[h!]
    \centering
\caption{Size of training datasets (No. of tweets)}
\label{tab:training}
    \centering
    \begin{tabular}{|l|l|l|l|l|l|}
    \toprule
        \textbf{Language} & \textbf{Aspect} & \textbf{Total} & \textbf{Positive} & \textbf{Neutral} & \textbf{Negative} \\ \midrule
        cs & Ukraine & 1638& 632& 447& 559
\\ 
        cs & Russia & 1722& 576& 275& 871
\\ \midrule
        pl & Ukraine & 640& 205& 263& 172
\\ 
        pl & Russia & 570& 202& 164& 204
\\ \midrule
        hu & Ukraine & 628& 202& 203& 223
\\ 
        hu & Russia & 556& 181& 145& 230
\\\botrule
    \end{tabular}
\end{table}

Each model in combination with each translation mode was then fine-tuned on each sub-dataset training split. For Llama2/3 and Mistral we used the PEFT (Parameter-Efficient Fine-Tuning) adapter-based technique\citep{liu2022few} using the Python PEFT library\footnote{https://huggingface.co/docs/peft}. The number of tuned parameters varied between 3.5--4 millions. The training was run for 10 epochs on all models. We regularly measured model's performance on the test set during the training, and all subsequent experiments were run on the model's best checkpoint.
Both training and inference was run on a server 2 x 2060 RTX (8GB) for smaller BERT-derived models, and another server with 2 x NVIDIA V100 (32GB) for larger models.

\subsection*{Inference}

All compared models were prompted the same way using the testing split of each annotated sub-dataset. We also used GPT4 as a reference model for inference, with no fine-tuning.
After a series of experiments with manually designed prompts in both English and target languages without conclusive results, we decided for a simple English prompt in all multilingual experiments:\\
\\
{\it reference tweet: \{ref\_tweet\}\\ 
tweet: \{tweet\}\\ 
The sentiment of the tweet towards \{aspect\} is\dots}\\

We also studied how the in-context learning \citep{wang2020generalizing} changes the performance of the models. The improvement was negligible in most cases except the case of GPT4. The full prompt is given in Appendix \ref{app_incontext}.


\section{Results}
\label{sec_experiments}

We conducted an extensive series of tweet sentiment classification experiments that varied in the following settings:
\begin{itemize}
    \item sentiment aspect (Russia/Ukraine)
    \item language of the tweet (CS/SK, HU, PL)
    \item language model (BERT, BERTweet, Llama2, Llama3, Mistral, GPT4)
    \item tweet translation (DeepL, Helsinki translator, none)
    \item positive/neutral/negative classification, or only positive/negative
    \item the presence of a reference tweet (to which the classified tweet reacted)
\end{itemize}

\begin{figure}
    \centering
    \includegraphics[width=\textwidth]{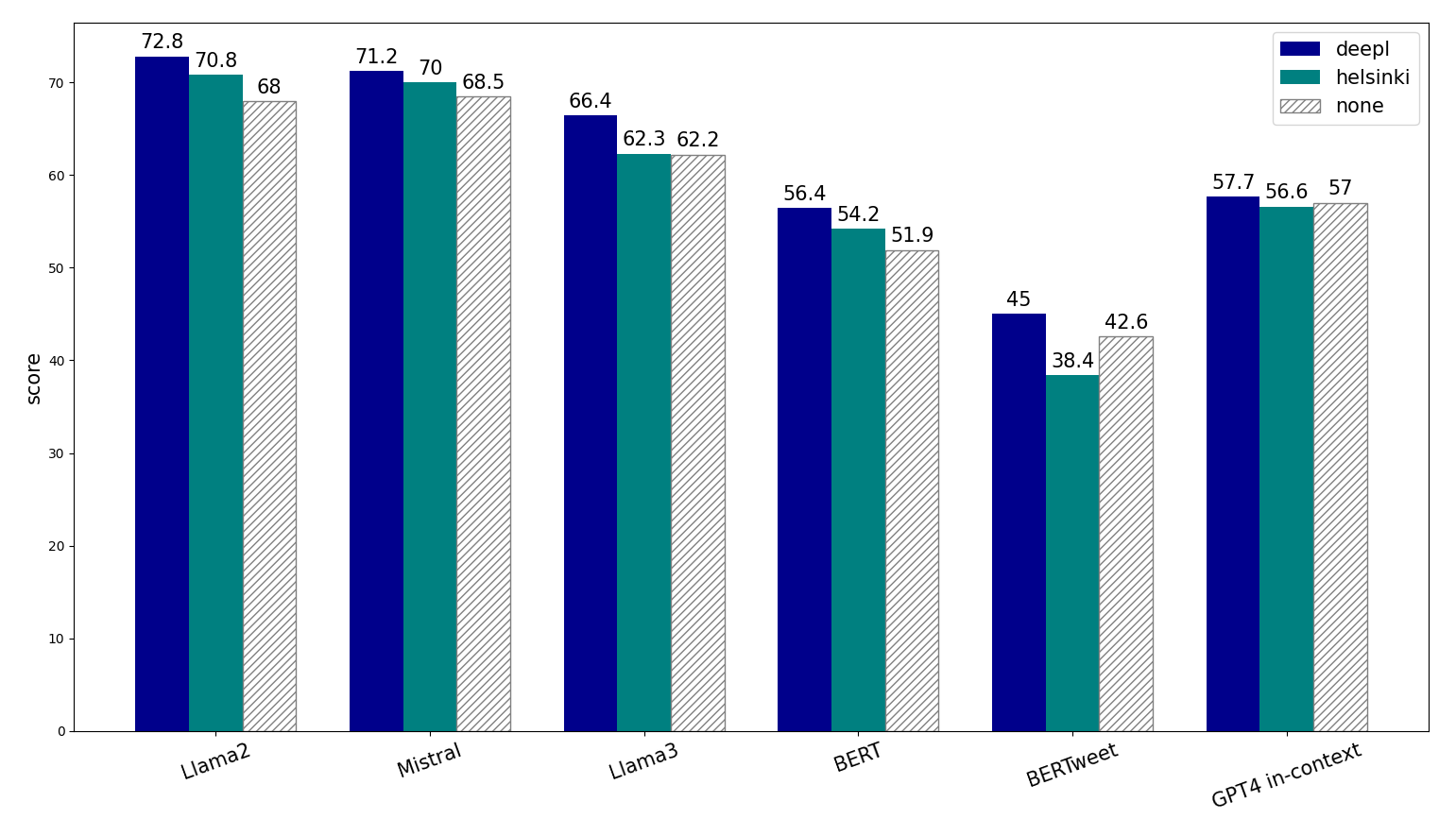}
    \caption{Macro-averaged F1-score by language models and translation, as listed in Table \ref{tab:avg-model-target-translator}.}
    \label{fig:avg-model-translator}
\end{figure}

The goal was to study the influence of individual settings (and their combinations) on the classification performance. We used standard metrics: accuracy and macro-averaged recall, precision and F1 score \citep{rainio2024evaluation}. Unless stated otherwise, we used three-valued (positive/neutral/negative) sentiment classification. After a few initial tests, all experiments were run without in-context learning, with the exception of the GPT4. Table \ref{tab:avg-model-target-translator} visualised at Fig. \ref{fig:avg-model-translator} summarises main results organised by language models and the type of translation. 

\begin{table}[h!]
    \centering
\caption{Macro-averaged F1-score by language models and translation, averaged over all languages, both aspects and (non)use of reference tweet, i.e., each score is an average of 12 experiments.}
\label{tab:avg-model-target-translator}
\begin{tabular}{c|cccccc}
\toprule
   &  Llama2&  Mistral &  Llama3 &  BERT &  BERTweet & GPT4\\
\midrule
DeepL&  72.8
&  71.2&  66.4&  56.4&  45.0& 57.7 \\
Helsinki&  
70.8&  70.0&  62.3&  54.2&  38.4& 56.6\\
None&  68.0&  68.5&  62.2&  51.9&  42.6& 57.0\\
\botrule
\end{tabular}
\end{table}

\subsubsection*{Comparison with the SOTA} 

The TweetEVAL leaderboard\footnote{\url{https://github.com/cardiffnlp/tweeteval}} introduced in \citep{barbieri2020tweeteval} marks TIMELM-21 as the SOTA model for sentiment classification with macro-averaged recall 73.7 for three-valued sentiment classification, closely followed by BERTweet with recall 73.4. Our best setting (LLama2, translation by DeepL, averaged over all aspects and languages) gave both recall and F1-score 72.8 (73.8 without the use of reference tweets). Our task is topically much narrower than TweetEVAL, on the one hand. On the other hand, TweetEVAl is monolingual and BERTweet was trained on 850M English tweets, while we fine-tuned our models using three underrepresented languages with small fine-tuning datasets (about 6K tweets in total).

Considering further the UMSAB multilingual Twitter benchmark\footnote{\url{https://github.com/cardiffnlp/xlm-t}} \citep{barbieri2022xlm}, the best reported model is XLM-Tw Multi with F1-score 69.4, macro-averaged over eight languages. Again, this task is wider than ours but XLM-Tw Multi used much larger fine-tuning dataset (198M multilingual tweets), therefore we cannot provide an exact comparison which may be the subject of future work.

\subsubsection*{Comparison of language models} 

Our best performing models were Llama2 and Mistral with almost equal average F1-scores. Surprisingly, Llama3 scored by approx. 6\% F1 worse than Llama2. It is all the more surprising since Llama3  was aware of the contemporary context  (actions and names of politicians, related regional events etc.) which is essential to understand the tweets. Llama3 could also potentially access in its pre-training period the same Twitter/X data we used for fine-tuning. Neither GPT4 (without fine-tuning but with in-context learning) performed well compared to Llama2 or Mistral.  Surprisingly, BERTweet large, the SOTA model according to the TweetEval \citep{barbieri2020tweeteval}, performed even worse than BERT base (Table \ref{tab:training}). 
It might be attributed to its size smaller than Llama2 or Mistral, plus it was pre-trained on data until 2020 without recent context. 

It seems that some models (Llama2, Mistral, BERT) strongly benefited from fine-tuning while some others (Llama3, BERTweet) were more ``stiff'' and less tunable. The proportion of PEFT-tuned parameters was slightly smaller in Llama3 than in Llama2 or Mistral (Table \ref{tab:models}). However, this does not correspond to differences in their performance.

\subsubsection*{Translation to English}

In the overwhelming majority of settings (see Tables \ref{tab:avg-model-target-translator}, \ref{tab:best-reduced} and \ref{tab:best-nonreducted}), all tested LLMs performed better with English-translated datasets (for both fine-tuning and testing), and the DeepL gave better results than the Helsinki translator. Therefore, despite successful multilingual models as XLM-R \citep{conneau2020unsupervised} or XLM-T \citep{barbieri2022xlm}, a good translation to English still provided an advantage.

\subsubsection*{Language and culture differences}

\begin{figure}
    \centering
    \includegraphics[width=\textwidth]{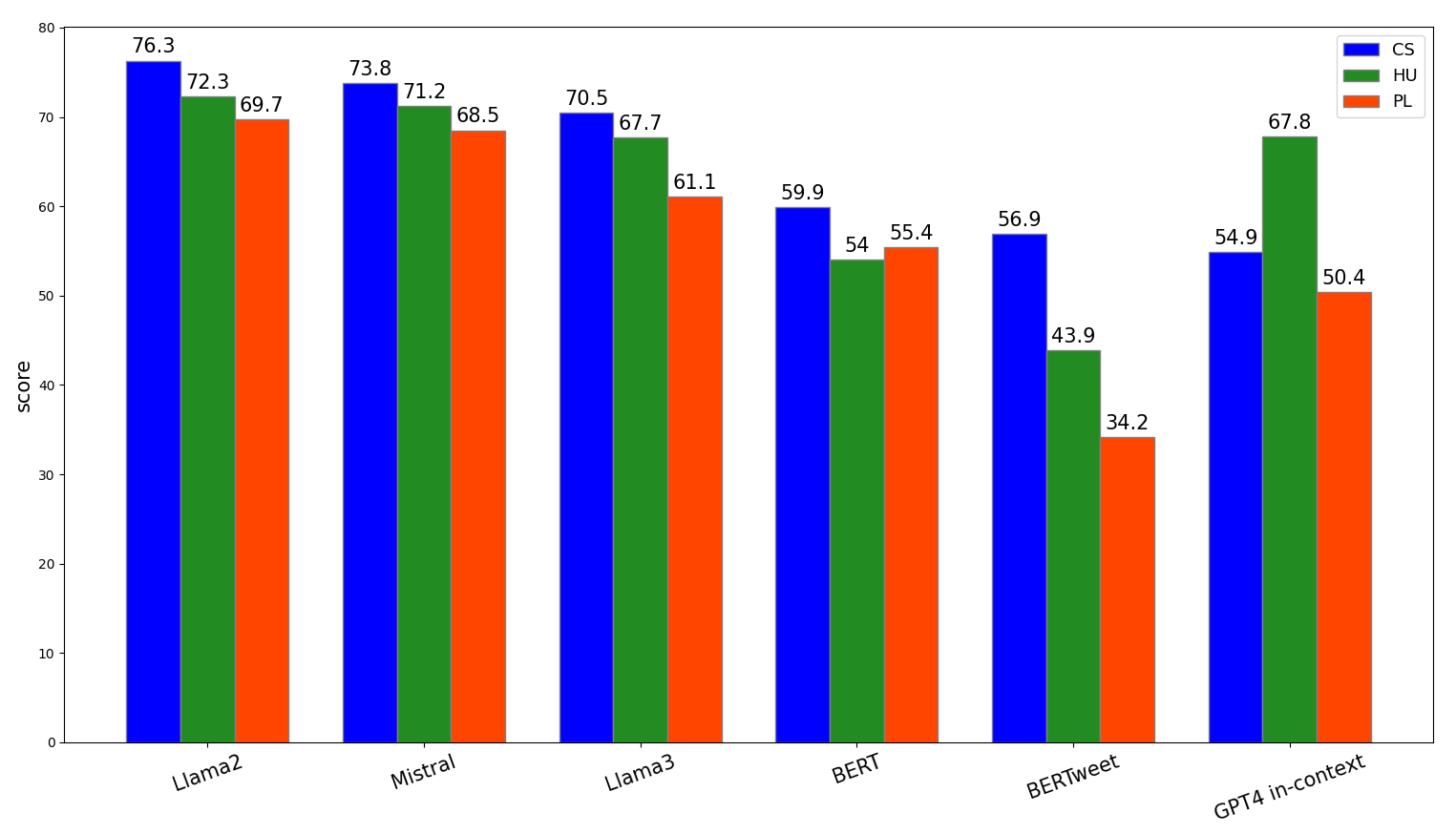}
    \caption{Macro-averaged F1-score by models and languages of tweets, averaged over all types of translation, both aspects and (non)use of the reference tweet, i.e., each score is an average of 12 experiments.}
    \label{fig:avg-model-lang}
\end{figure}

Figure \ref{fig:avg-model-lang} summarises experiments on individual languages using the DeepL translation.  Rather surprisingly, almost all models performed poorer for the Polish language. These results correlate neither to the support of fine-tuning sets of individual languages which was almost the same for PL and HU (see Table \ref{tab:training}) nor to the type of translation (the results were similar for untranslated tweets), and they cannot be attributed to pre-training either (e.g., GPT4 gave very good results for Polish on the multilingual MMLU benchmark  \citep{GPT4-report}).  Note that also vanilla models (LLama2, LLama3, GPT4) put Polish tweets at a disadvantage compared to other languages (Table \ref{tab:vanilla_llama}). 
A detailed analysis showed that many positive Polish tweets (either towards Ukraine or Russia) were classified as negative by the models. These tweets contained more complex thoughts reflecting the fact that Poland was historically more interconnected with Ukraine than CS/SK or HU. A few examples of misclassified tweets can be found in Appendix \ref{app_misclassified}. 

\begin{figure}
    \centering
    \includegraphics[width=\textwidth]{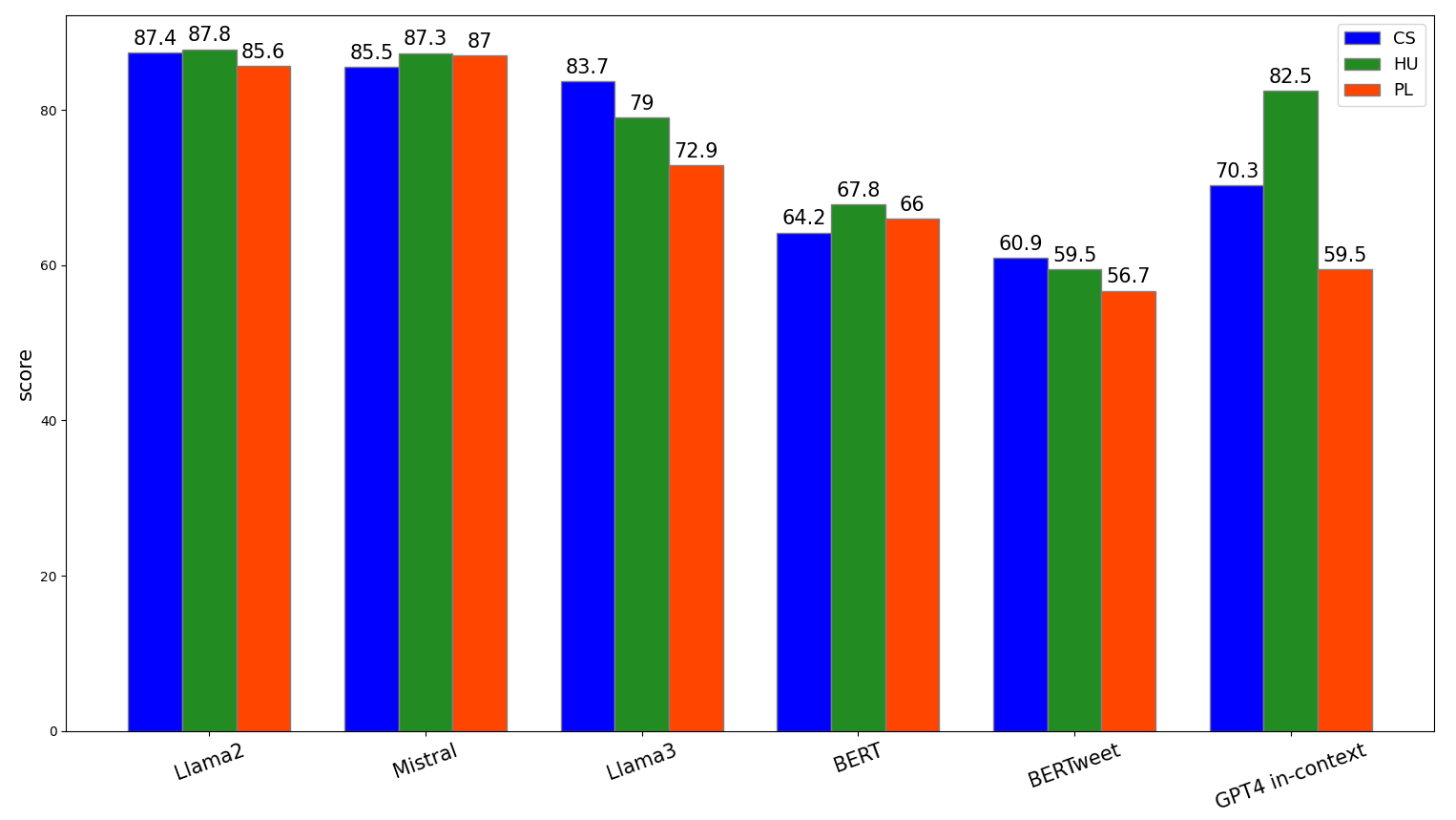}
    \caption{Macro-averaged F1-score by language models and translation for \emph{two-valued classification (positive/negative)}, averaged over all types of translation, both aspects and (non)use of the reference tweet, i.e., each score is an average of 12 experiments}
    \label{fig:avg-model-lang-reduced}
\end{figure}

However, when repeating fine-tuning with two-valued sentiment classification (Fig. \ref{fig:avg-model-lang-reduced},  some models (Llama2, Mistral, BERT) almost erased differences between languages. This supports our hypothesis that these models are better tunable than the others. We also relate this observation to the sizes of the training sets discussed bellow.

\subsubsection*{Fine-tuning vs. in-context learning}

To get a comparison, we tested three vanilla models: LLama2, Llama3 and GPT4. Fine-tuning improved dramatically the performance of Llama2/3 over vanilla models (compare Fig. \ref{fig:avg-model-lang} and Tab. \ref{tab:vanilla_llama}). In contrast, we tried also in-context learning which did not matter much for most tested models except GPT4 for which it improved the accuracy on all datasets by 5--10\%. We therefore confirm the observations of \citep{liu2022few} about prevalence of fine-tuning over in-context learning.

\begin{table}[h!]
    \centering
\caption{Macro-averaged F1-score of vanilla Llama2, Llama3 and GPT4 for individual languages and targets, using the DeepL translation.}
\label{tab:vanilla_llama}
\begin{tabular}{llrrr}
\toprule
 Language & Target & Llama2  &Llama3 &GPT4\\
\midrule
  cs& ua& 38.5 & 37.8 &48.8 \\
  cs& ru& 40.2 & 47.1 &49.8 \\
\midrule
  hu& ua& 41.7 & 44.7 &55.9 \\
  hu& ru& 53.9 & 47.4 &60.5 \\
\midrule
 pl& ua& 33.7 & 24.3 &39.8 \\
  pl& ru& 34.8& 35.7&46.1\\
\botrule
\end{tabular}
\end{table}


\subsubsection*{Size of the training sets}
The support of the CZ/SK training set was approximately three times that of HU or PL which were almost equal, see Table \ref{tab:training}.  This discrepancy allowed for some interesting observations. In the simpler task of two-valued classification, almost all fine-tuned models returned scores irrelevant to the language, implying that the training sets with about 800 tweets were sufficient to bridge the language differences. In the case  of three-valued classification, however, CZ/SK tweets were preferred in all models except GPT4. This indicates that for this harder task the training set of some 1200 tweets was insufficient and the models fine-tuned better with CZ/SK set of size approx. 3400.


\subsubsection*{Reference tweet usage}
Presence of the reference tweet during training and testing did not make any significant difference and, actually, it worsened the scores by 1--2\% in most cases. This might result from the fact that the relation between tweet and its reference tweet is often vague, they might either support or contradict each other. 

\subsubsection*{Model and human bias}  

In the framework of the current situation where Russia is described as the aggressor, human annotators who know more about the context may tend to see the situation in terms of cause and effect, and therefore their sentiment determination may be biased differently than the models.  In particular, LLMs struggled with tweets neutral (or positive) to a given aspect, but generally negative - for example, addressing bombing, war, attack. Well performing models as Llama2 or Mistral usually showed significantly lower precision and recall for the neutral class than for the negative or positive one.


\subsubsection*{Aspect of sentiment}
Fine-tuned models did not show significant differences in scores between aspect (Russia/Ukraine) in the case of two-valued sentiment: usually the tweets targeting Russia were predicted by approx 2\% better, on average, than those targeting Ukraine. Interestingly, BERT was the only model scoring better for Ukraine tweets. In the case of three-valued sentiment, the differences rose up to 4\% for Llama and Mistral models, and up to 10\% for the other models (supporting data can be found in the supplementary material).  The same applied also to vanilla models, see Table \ref{tab:vanilla_llama}. For a few examples see Tables \ref{tab:best-reduced} and \ref{tab:best-nonreducted}. 
As already mentioned above, models tended to classify the general sentiment of the tweet (which was often negative) rather than the sentiment towards a given aspect. Therefore, they misclassified some tweets positive about Ukraine which were simultaneously negative about war or Russia.

\subsubsection*{Scores for individual classes}
All experiments in Section \ref{sec_experiments} used macro-averaged recall, precision and F1 scores since mostly the scores were similar for all classes (negative/neutral/positive), with a few exceptions.
Notably in Hungarian, recall of positive class was often by approx. 10\% lower than that of the negative class, and the trend was opposite in precision, meaning that the models tended to classify Hungarian tweets more negative than human annotators. 


\section{Conclusion}
\label{sec_conclusion}
We addressed the problem of fine-tuning large language models for the task of aspect-based classification of sentiment towards a given aspects. For our experiments we chose the theme of the Russian-Ukrainian conflict which is still ongoing and it polarises society in a strong way. We narrowed our dataset to the V4 (Czech Republic, Slovakia, Poland, Hungary) language space, chosen due to its  proximity to the conflict, and also because we are not aware of a similar study based on this underrepresented group of languages.
Data were collected using the academic API from the social network Twitter (X) during the first half of 2023, and split into three independent language-specific datasets which were manually annotated.

 We performed fine-tuning of selected models (LLama2, LLama3, BERT, Bertweet, Mistral) separately for each dataset in several variants, using either the original language or an English translation with the Helsinki or DeepL translator, and focusing on sentiment towards a specific entity (Russia, Ukraine). Further detailed settings are described in Section \ref{sec_methods}. 
The results were evaluated using basic metrics - accuracy and macro-averaged precision, recall and F1.  Furthermore, the GPT4 (with or without in-context learning) was used as a reference model.
The best performing (SOTA-level) model was the Llama 2 followed closely by Mistral, both in combination with the DeepL translation. Rather surprisingly, Llama3, BERTweet and GPT4 performed significantly worse.

 Results of the experiments revealed several interesting phenomena related to differences between models and between languages. One conclusion is  that the fine-tuning, even with as few as hundreds of samples, was able to draw the model's attention to the desired aspects and also to balance language and culture differences (at least for some models). The results also indicate that the success of fine-tuning is highly model- and task-dependent, as found also in other studies such as \citep{zhang2024scaling}. 

\subsubsection*{Future work}

 The classification of  sentiment in tweets (or similar short texts) can suffer from various sources of bias, such as the subconscious assumptions of cause-and-effect, the lack of contemporary context or human versus model bias towards a certain aspect. To understand and evaluate biases in models towards specific ongoing polarising themes, we propose to create a synthetic dataset focused on Russo-Ukraine conflict. We would study various topics such as general sentiment of sentences, aspect based sentiment, sentiment of the implication etc. on several models in different setting using a more granular approach.

Another recent trend is the distillation of knowledge to smaller models (such as Microsoft Phi-3 \citep{abdin2024phi} or 1-bit models \citep{ma2024era}) to make inference on narrow tasks cheaper and also to avoid the necessity of manual annotation. We have already seen in our experiments that fine-tuned medium-sized models easily surpass universal large ones as GPT4.
 
Working with different languages and information spaces and the use of different methods requires integration of both methods and acquired experience to a more general platform dealing with public information in cyberspace. Such a platform would contain components that take care of the collection, storage, annotation, analysis and downstream usage of data from heterogeneous sources (social networks, open internet). These components would be anchored by a web interface through which the data would be accessed in a unified and controlled manner \citep{zrec2, zrec1}.

\backmatter
\bmhead{Supplementary information}
Supplementary data are available under \\
\url{https://github.com/zrecorg/zrec-paper-a-study-on-eastern-european-v4-languages}
%

\bmhead{Acknowledgements}
The authors thank to Ludmila Rezkov\'a for her help with data labelling and for creating the charts.

\bmhead{Funding}
This work was supported by the Silesian University in Opava under the Student Funding Plan, project SGS/9/2024.

\section*{Declarations}
\noindent
\textbf{Conflict of interest} The authors declare that they have no conflict of interest.

\bibliography{references}

\newpage
\begin{appendices}

\section{Selected experimental results}
\label{app_tables}
The following two tables compare 10 top-scoring experiments of fine-tuned models with negative/positive classification, and negative/neutral/positive classification, providing detailed metrics. 

\begin{table}[h!]
    \centering
\caption{Top 10 macro-averaged F1-scores among all experimental settings with negative/positive classification}
\label{tab:best-reduced}
\begin{tabular}{llllllllll}
\toprule
 & aspect & lang.  & model & translator & ref. & prec.& recall& F1& acc.\\
 &        &       &       &            & tweet &      &    &      &  \\
\midrule
1
& ru& hu& llama2
& deepl
& yes
& 92.7& 91.6& 92& 92.2
\\
2
& ua& hu& mistral
& deepl
& yes
& 91.8& 91.4& 91.5& 91.6
\\
3
& ru& hu& llama2
& deepl
& no
& 90.7& 89.6& 90& 90.3
\\
4
& ru& pl& llama2
& helsinki
& no
& 89.4& 89.2& 89.2& 89.2
\\
5
& ua& pl& mistral
& deepl
& no
& 88.3& 88.4& 88.3& 88.4
\\
6
& ru& pl& llama2
& helsinki
& yes
& 88.2& 88.2& 88.2& 88.2
\\
7
& ru& pl& mistral
& deepl
& no
& 88.2& 88.2& 88.2& 88.2
\\
8
& ru& cs& mistral
& deepl
& no
& 87.8& 88.2& 88& 88.4
\\
9
& ru& cs& llama2
& deepl
& no
& 88.8& 87.4& 88& 88.7
\\
10
& ua& cs& llama2& deepl& no& 87.9& 87.9& 87.9& 87.9\\
\botrule
\end{tabular}
\end{table}


\begin{table}[h!]
    \centering
\caption{Top 10 macro-averaged F1-scores among all experimental settings with negative/neutral/positive classification.}
\label{tab:best-nonreducted}
\begin{tabular}{llllllllll}
\toprule
& aspect & lang. &model & translator & ref. & prec.& recall& F1&acc.\\
 &        &      &&            & tweet &      &    &      &\\
\midrule
1 
& ru& cs&llama2
& deepl & no
& 80.7& 79.2& 79.9&82.4
\\
2 
& ua& cs&llama2
& deepl & no
& 76.7& 77.7& 76.9&76.8
\\
3 
& ru& cs&mistral
& deepl & no
& 76.6& 75.5& 76&79.6
\\
4 
& ru& hu&mistral
& deepl & no
& 75.7& 76.4& 75.8&76.3
\\
5 
& ru& cs&llama2
& deepl & yes
& 75.4& 75.5& 75.2&78.7
\\
6 
& ru& cs&mistral
& deepl & yes
& 76.4& 73.6& 74.8&78.7
\\
7 
& ru& hu&llama2
& deepl & no
& 74.5& 74.9& 74.6&75.5
\\
8 
& ru& hu&llama3
& deepl & yes
& 75.7& 74.3& 74.3&76.3
\\
9 & ru& hu&llama2
& helsinki & yes
& 74.5& 73.4& 73.7&75.5
\\
10& ua& hu&llama2& helsinki & yes& 74.2& 73.5& 73.6&73.9\\
\botrule
\end{tabular}
\end{table}

\section{Prompt for in-context learning}
\label{app_incontext}
{\it 
In 2022, the war between Russia and Ukraine began. There is an information war between both sides of the conflict. Twitter users with a pro-Russian opinion and users with an anti-Russian opinion accuse each other. Anti-Russian users often label Russia as fascist and Putin as the culprit of the war. Pro-Russian users often claim that Russia is defending itself against the Ukrainian fascists or that the western countries are responsible for the war.\\
\\ 
You will receive a tweet and its reference tweet as input. Classify the sentiment in tweet regarding to Russia. The sentiment must be related to tweet not to reference tweet. Sentiment can be positive, negative or neutral. Do not append any justification beyond sentiment label.
}

\section{Examples of misclassified tweets}
\label{app_misclassified}
\begin{table}[h!]
\centering
\begin{tabular}{|l|l|p{7cm}|l|l|}
\toprule
\textbf{Model} &  \textbf{Aspect}& \textbf{Tweet} &                                                                                            Label&Predict.\\ 
\midrule
LLama3&  Ukraine& The West will put itself at colossal risk if it delivers F-16 fighter jets to Ukraine              & neutral&positive
\\ \midrule
LLama3&  Ukraine& Ukraine is, was and will be Russian territory. Russians are liquidating an internal enemy on their territory. This is how 80\% of the world's population perceives the conflict  & negative&positive
\\ \midrule
 LLama3& Russia& @Just why should Russia attack Poland? Purpose: profit? For fun?& positive&negative\\ \midrule
LLama2&  Ukraine& Czech TV is disinformation shit. I haven't seen that shit in six months and I'm better. ( See. Ukraine is fighting back, Ukraine is winning, Ukraine is not ruled by a fascist junta, Putin is dead, Fiala is dignified and shit like that!!!!)& negative&positive\\ 
\midrule
LLama2&  Ukraine&It's just a sliver. Back in 2019 Putin publicly declared that Ukrainians were committing unpublishable atrocities. Considered propaganda, I didn't believe it. I know that schools read Russian, so I "looked around". It's really absolutely appalling. Nothing will be forgotten.  & negative&positive\\
\midrule
 LLama2& Russia& Unfortunately, in every country we have a part of the hate society: Sweden 20\%, the US 45\%, Russia 80\% and in our country 30\%. Therefore, it is necessary to introduce regulations to limit the freedom of action of fascist and populist groups.& neutral&negative\\ 
 \midrule
GPT4 &  Russia& It will probably be that they would love this, because the goal and purpose of the West with the US in the lead has been the same for 78 years since the end of 2. SV, and it will destroy Russia, separate it into individual states and plunder its raw materials. The Russians already know and resist.                          & positive&negative\\ \botrule
\end{tabular}
\end{table}

\end{appendices}

\end{document}